\documentclass[sigconf]{aamas}

\usepackage{balance} 

\usepackage{nicefrac}
\usepackage{bm}
\usepackage{amsmath}
\usepackage{amsthm}
\usepackage{xcolor}
\usepackage{booktabs}
\usepackage[inline]{enumitem}
\usepackage[subpreambles=true]{standalone}
\usepackage{mystyle}
\usepackage{multirow}
\usepackage{algorithm}
\usepackage{algorithmic}
\usepackage{newtxtext,newtxmath}
\usepackage{soul}

\setcopyright{ifaamas}
\acmConference[AAMAS '26]{Proc.\@ of the 25th International Conference
on Autonomous Agents and Multiagent Systems (AAMAS 2026)}{May 25 -- 29, 2026}
{Paphos, Cyprus}{C.~Amato, L.~Dennis, V.~Mascardi, J.~Thangarajah (eds.)}
\copyrightyear{2026}
\acmYear{2026}
\acmDOI{}
\acmPrice{}
\acmISBN{}

\newcommand{\anonimity}{\url{https://github.com/NathanGavenski/UfO}}

\acmSubmissionID{294}

\title{Towards Generalisable Imitation Learning Through Conditioned Transition Estimation and Online Behaviour Alignment}

\author{Nathan Gavenski}
\affiliation{
  \institution{King's College London}
  \city{London}
  \country{United Kingdom}}
\email{nathan.schneider_gavenski@kcl.ac.uk}

\author{Matteo Leonetti}
\affiliation{
  \institution{King's College London}
  \city{London}
  \country{United Kingdom}}
\email{matteo.leonetti@kcl.ac.uk}

\author{Odinaldo Rodrigues}
\affiliation{
  \institution{King's College London}
  \city{London}
  \country{United Kingdom}}
\email{odinaldo.rodrigues@kcl.ac.uk}

\begin{abstract}
    State-of-the-art imitation learning from observation methods (ILfO) have recently made significant progress, but they still have some limitations: they need action-based supervised optimisation, assume that states have a single optimal action, and tend to apply teacher actions without full consideration of the actual environment state.
    While the truth may be out there in observed trajectories, existing methods struggle to extract it without supervision.
    In this work, we propose \method (\abbrev) that addresses all of these limitations.
    \abbrev learns a policy through a two-stage process, in which the agent first obtains an approximation of the teacher's true actions in the observed state transitions, and then refines the learned policy further by adjusting agent trajectories to closely align them with the teacher's.
    Experiments we conducted in five widely used environments show that \abbrev not only outperforms the teacher and all other ILfO methods but also displays the smallest standard deviation. 
    This reduction in standard deviation indicates better generalisation in unseen scenarios.
\end{abstract}


\newcommand{\BibTeX}{\rm B\kern-.05em{\sc i\kern-.025em b}\kern-.08em\TeX}

\begin{document}


\pagestyle{fancy}
\fancyhead{}


\maketitle 


\section{Introduction} \label{sec:introduction}

One way to learn how to perform a task is to observe someone else execute it and then attempt to imitate their actions.
In computer science, this is called \textit{imitation learning} (IL), where an agent observes a ``teacher''.
Unlike other learning approaches, IL leverages the teacher's guidance, thus avoiding blindly learning by trial-and-error.
 
\textit{Learning from demonstration} (ILfD) is the most common approach for IL, where the agent uses sequences of states and actions performed by the teacher.
In practice, it suffers from two significant drawbacks: poor generalisation in environments with multiple solutions and the scarcity of action-labelled data~\cite{gavenski2022how}.
\textit{Learning from observation} (ILfO) mitigates the latter by learning a task with access to only sequences of states via self-supervision.
This allows agents to use much more easily obtainable data, such as video recordings of game-playing.

ILfO typically employs one of two approaches: \textit{behavioural cloning} (BC) or \textit{adversarial imitation learning} (AIL).
BC seeks to match a teacher's actions by minimising a supervised loss, while AIL aims to make the agent's behaviour indistinguishable from that of the teacher.
To avoid ILfD's generalisation problem, BC assumes each state in the environment can be reached by only one optimal action, which is not always the case. 
To address this and correct the agent's behaviour, AIL collects samples via \textit{online play} (direct environment interaction).
Yet, these strategies still have shortcomings.
BC tends to overfit, and AIL can fall into \textit{behaviour-seeking} mode (blindly mimicking the teacher's behaviour, irrespective of the actual environment state). 
This leads to poor performance in environments that require the underlying action intent or the causal relationships between actions and outcomes for success.

To tackle these shortcomings, recent works~\cite{gavenski2024explorative,monteiro2023self} use self-supervision to approximate the teacher's behaviour, with others employing an artificial reward function~\cite{zhu2020opolo,li2023mahalo}.
However, these works still assume that states can be reached only by a single action, rely on an action-based loss function, and exhibit behaviour-seeking due to reliance on the quality and coverage of the data samples. 

In this work, we introduce \method{} (\abbrev{}), a novel unsupervised approach to imitation learning from observation.
Unlike prior ILfO methods~\cite{gavenski2024explorative,monteiro2023self,li2023mahalo,torabi2018gaifo}, which rely on supervised losses to infer or reconstruct teacher actions, \abbrev{} learns without action-based supervision.
Instead, it derives its understanding of behaviour from the causal structure of state transitions.
The main contribution of \abbrev{} is the reconstruction stage, where a policy and a generative model are jointly trained to approximate the environment’s dynamics through mutual optimisation: the policy seeks actions that best condition the generative model to reproduce the teacher’s transitions, and the generative model refines its predictions based on environment transitions.
This unsupervised coupling eliminates dependence on action-level supervision while constraining learning to behaviourally meaningful transitions, mitigating the behavioural-seeking bias common in ILfO.
A subsequent adversarial refinement stage further improves policy generalisation through online play, enabling \abbrev{} to surpass the teacher’s performance and outperform all competing methods across five benchmark environments.
Together, these results establish \abbrev{} as the first method to achieve effective unsupervised imitation purely from observation, without relying on supervised action inference or external rewards.
\section{Background} \label{sec:background}

In this work, we assume the environment to be a Markov Decision Process (MDP) $\mdp = \tuple{\MDPstate, \action, \transition, r, \gamma}$, in which $\MDPstate$ is the state space, $\action$ is the action space, $\transition$ is a transition function $\transition: \MDPstate \times \action \rightarrow \MDPstate$, $r$ is the immediate reward function, and $\gamma$ is the discount factor~\cite[Ch. 3]{sutton2018reinforcement}.

Solving an MDP yields a policy $\policy$ with a probability distribution $P$ over actions. 
In each state $s$, $P(a \mid s)$ is the probability of taking action $a$ in $s$.
For simplicity, we assume that $\policy(s)$ will determine the action the agent takes in state $s$, i.e., the action $a$ such that $P(a \mid s)$ is the highest. 
We will differentiate policies via appropriate subscripts. 

The objective of IL is to generate an agent policy $\agent$ that approximates the teacher's policy $\teacher$ from a set of `trajectories' representing the behaviour of the teacher solving the task. Trajectories may consist of sequences of state-action pairs or simply sequences of state transitions. We will use the symbol $\Tau$ to denote the set of all trajectories, irrespective of their type. Notice that IL does not assume access to the teacher's reward function or discount factor.

If the trajectories include the actions performed by the teacher, imitation learning becomes a supervised learning problem, known as \textit{behavioural cloning} (BC), optimising:
\begin{equation} \label{eq:bc}
    \argmin_\agentW \textstyle\sum_{\tau \in \Tau} \sum_{(s,a) \in \tau} \Loss(a, \agent(s)).
\end{equation}
Above, we minimise the supervised loss function $\Loss$ (such as the mean square error) between the action $a$ the teacher chooses at state $s$ (annotated in a trajectory $\tau$) and the action $\agent(s)$ the agent policy thinks it should take.
Yet, relying on annotated actions from a teacher might not be feasible.
In some cases, such as when the teacher's data consists of video recordings, it might be impossible to retrieve which action was actually taken.
Therefore, ILfO assumes that trajectories contain only sequences of states and generate agent policies based solely on these sequences.
The two most common ILfO approaches are behavioural cloning and adversarial imitation learning (see~\citet{gavenski2024imitation} for details).

\subsection{Behavioural Cloning Methods}

Behavioural Cloning for ILfO tries to generate the agent policy using \textit{inverse dynamics models} (IDM) and \textit{forward dynamics models} (FDM).
IDM tries to predict the action responsible for a transition ($s,s'$),
where FDM tries to model $\transition$ directly.
The disadvantage of these dynamic models is that although the MDP consists of a single function $\transition$, they rely on learning one of many possible transition functions $\transition' \subseteq \transition$.
IL methods that rely on IDM~\cite{gavenski2024explorative,monteiro2023self} use teacher's trajectories $\tau_\teacherW$ to create \textit{self-supervised} actions $\tilde{a} \in \{\text{IDM}(s, s') \mid (s, s') \in \tau_\teacherW \}$ to train their policies using Eq.~\ref{eq:bc}.
On the other hand, methods that rely on FDM use the teacher's trajectories coupled with action~\cite{edwards2019ilpo} or temporal~\cite{paolillo2023dynamical} information to condition the next state generation and optimise the agent using:

\begin{equation}
    \argmin_{\agentW} \argmin_{\generatorW} \textstyle\sum_{\tau \in \Tau} \sum_{(s, s') \in \tau} 
    \Loss(s', \generator(s, \agent(s))),
\end{equation}

\noindent
where $\generator$ is a generative model approximating $\transition$, and $\Loss$ is still a supervised loss function used to minimise the loss between the actions chosen by the teacher and agent. 

These methods rely on the assumption that the teacher transition function is an \textit{injective function}, i.e., $\transition(s_i, a_i) \neq \transition(s_j, a_j)$, whenever $(s_i, a_i)\neq (s_j, a_j)$.
However, this rarely holds in practice, resulting in the learning process encouraging the agent to favour a single action during a transition (even when multiple alternatives are viable).
Hence, the agent tends to learn disproportionately skewed policies.
The inability to give the correct weights to action alternatives in unseen scenarios often leads to failure~\cite{swamy2021gap}. 
We will show how to tackle this issue in Section~\ref{sec:sub:reconstruction_step}.

\subsection{Adversarial Imitation Learning Methods}

Adversarial imitation learning employs a \textit{discriminator} that attempts to distinguish between the actions chosen by the agent and those chosen by the teacher for states encountered during online play.
Instead of learning an action mapping function similar to BC, it tries to approximate the teacher policy by how the agent behaves.
AIL can be further divided into two categories: (i) \textit{inverse reinforcement learning}, and (ii) \textit{generative adversarial learning}.

Most AIL approaches learn by minimising a variation of the adversarial loss from \citeauthor{goodfellow2014generative}~\shortcite{goodfellow2014generative}:
\begin{equation} \label{eq:ail}
    \begin{split}
        \argmin_\discriminatorW \argmax_\agentW\! \sum_{\tau \in \Tau}\! \sum_{s \in \tau}
            log(\discriminator\!\left( s, \teacher\!\left( s \right)\right)\!+
            log(1\!\!-\!\!\discriminator\!\left(s, \agent\!\left( s \right)\right)
    \end{split}
\end{equation}
\noindent
where $\discriminator$ is a \textit{discriminator model} that identifies whether a pair ($s, a$) is from the teacher or from the agent.
It is important to note that Eq.~\ref{eq:ail} also does not assume access to $\teacher$.
It only requires the agent's policy to follow the behaviour of the teacher's.
Therefore, to ``fool'' the discriminator, the agent attempts to follow the teacher's trajectories as closely as possible.
For inverse reinforcement learning approaches, $\discriminator$ trains to predict a value between $\left[0, 1\right]$, where $0$ means the model thinks $(s,a)$ is from the agent and $1$ means it is from the teacher.
Generative adversarial learning methods use the gradient from $\discriminator$ to teach the policy how to behave as the teacher.
For ILfO methods, Eq.~\ref{eq:ail} uses state transitions $(s', s)$ rather than state-action pairs to discriminate.

Dependence on adversarial losses requires IL approaches to rely heavily on the teacher samples' coverage and quality.
By forcing the agent to behave precisely as its teacher, adversarial methods often fall into behaviour-seeking mode~\cite{zhu2020opolo} --- trying to emulate the teacher's behaviour too closely and forcing a trajectory that might not be appropriate in the current state of the environment.
To make things worse, IL methods attempt to reduce the number of teacher samples to lower the learning cost (for greater \textit{efficiency}), amplifying this issue further.
Section~\ref{sec:sub:adversarial_step} shows how \abbrev employs AIL sparingly to refine the agent's policy without causing it to blindly mimic the teacher's behaviour.
\section{Unsupervised IL from Observation} \label{sec:method}

In this work, we propose \method{}, a novel IL technique that avoids the need for action-based supervised optimisation, the reliance on the assumption of injective transition functions and the behaviour-seeking mode.
\abbrev{} is an unsupervised imitation learning method composed of:
\begin{enumerate*}[label=(\roman*)]
    \item a policy model $\agent$ to learn to predict the best action given a state $s$; 
    \item a \textit{conditioned} generative model $\generator$ to generate the next state $\generator(s, \agent(s))$, given a state $s$ and the action $\agent(s)$; and 
    \item a discriminator model $\discriminator$ to discriminate between agent and teacher trajectories $\discriminator(\tau_\agent \vert\vert \tau_\teacher)$.
\end{enumerate*}
Algorithm~\ref{algo:method} provides an overview of \abbrev's learning process.
\abbrev{} initialises all models with random weights and collects teacher trajectories $\Tau_\teacher$ (Lines~\ref{algo:line:initialise_parameters} and \ref{algo:line:collect_teacher}).
The learning process is divided into two stages.
In the reconstruction stage (Lines~\ref{algo:line:step1_begin}-\ref{algo:line:step1_end}), \abbrev{} learns how to approximate the teacher's policy from the teacher observations collected, whilst learning an approximation of the MDP's transition function (removing the need for supervised losses).
In the subsequent adversarial stage (Lines~\ref{algo:line:step2_begin}-\ref{algo:line:step2_end}), \abbrev{} refines the policy to more closely resemble the teacher's.

\begin{algorithm}[b!]
    \caption{\abbrev{}}
    \label{algo:method}
    \begin{algorithmic}[1] 
        \STATE Initialise parameters $\agentW$ for $\agent$; $\generatorW$ for $\generator$; and $\discriminatorW$ for $\discriminator$ \label{algo:line:initialise_parameters}

        \STATE Collect teacher $\teacher$ trajectories $\Tau_\teacher$ \label{algo:line:collect_teacher}
        \STATE Let $\bm{T}_\teacher = \{(s,s') \in \tau \mid \tau \in \Tau_\teacher\}$
        
        \STATE \textbf{Stage 1: Reconstruction} 
        \FOR {$i \gets 1$ to $\text{epochs}_r$} \label{algo:line:step1_begin}
            \FORALL {$(s, s') \in \bm{T}_\teacher$} \label{algo:line:step1_policy_begin}
                \STATE Update $\agentW$ with $\Loss_R \left( s', \generator \left( s, \agent \left( s \right)\right) \right)$
            \ENDFOR \label{algo:line:step1_policy_end}
            \STATE Collect agent $\agent$ trajectories $\Tau_\agent$ in the environment \label{algo:line:collect_agent_step1}
            \STATE Let $\bm{T}_\agent = \{(s, s') \in \tau \mid \tau \in \Tau_\agent\}$
            \FORALL {$(s, s') \in \bm{T}_\agent$} \label{algo:line:step1_generator_begin}
                \STATE Update $\generatorW$ with $\Loss_R \left( s', \generator \left( s, \agent \left( s \right)\right) \right)$
            \ENDFOR \label{algo:line:step1_generator_end}
        \ENDFOR \label{algo:line:step1_end}
       
        \STATE \textbf{Stage 2: Adversarial} 
        \FOR {$i \gets 1$ to $\text{epochs}_a$} \label{algo:line:step2_begin}
            \STATE Collect agent $\agent$ trajectories $\Tau_\agent$ in the environment \label{algo:line:collect_agent_step2}
            \FORALL {$\tau \in \Tau^\Delta_\agent \cup \Tau^\Delta_\teacher$}
                \STATE $y \gets \{1 \mid \tau \in \Tau_\teacher\} \cup \{0 \mid \tau \in \Tau_\agent\}$
                \STATE Update $\discriminatorW$ and $\agentW$ with $\Loss_D \left(y, \discriminator \left( \tau \right)\right)$
            \ENDFOR
        \ENDFOR \label{algo:line:step2_end}
    \end{algorithmic}
\end{algorithm}
\begin{figure*}
    \centering
    \includestandalone[width=\linewidth]{content/figures/tikz/method}
    \caption{Two-stage training cycle. (a) Each model is iteratively frozen and trained using samples from the teacher $\Tau_\teacher$ and the agent, respectively. (b) $\generator$ is frozen, and $\agent$ and $\discriminator$ are trained with agent samples from the environment and the teacher's.
    }
    \Description[Two-stage training cycle diagram]{
        Two-panel diagram showing iterative training stages. Panel (a) shows the reconstruction stage, in which the policy and discriminator are trained with the frozen generator using teacher trajectories. Panel (b) shows the adversarial stage where the policy and discriminator are trained with the generator frozen using agent samples from the environment and teacher demonstrations.
    }
    \label{fig:training_cycle}
\end{figure*}

\subsection{Reconstruction Stage} \label{sec:sub:reconstruction_step}

The reconstruction stage (illustrated in Fig.~\ref{fig:training_cycle}a) trains both generator and policy based on the teacher transition function and the agent transition function $\Tau_\agent$.
It does so, by iteratively \textit{freezing} $\generator$ and $\agent$, and training each model by using $\Tau_\agent$ and $\Tau_\teacher$, respectively.
This part of the training runs for $\text{epochs}_r$, which is significantly higher than the second stage.

For the policy portion of the training (Lines~\ref{algo:line:step1_policy_begin}-\ref{algo:line:step1_policy_end}), \abbrev{} uses the teacher samples to learn how to properly condition the generator based on the teacher's transitions following:
\begin{equation} \label{eq:lr_policy}
    \argmin_\agentW \textstyle\sum_{\tau \in \Tau_\teacher} \sum_{(s, s') \in \tau} \mid s' - \generator\left(s, \agent\left(s\right)\right)\mid.
\end{equation}
This part aims to enable the policy to learn how to accurately predict an action that allows the generator to determine the teacher's next state based on its learned transition function.
In other words, given that $\generator$ learns a sufficiently approximate transition function $T'\subseteq T$, if the agent predicts the same action that the teacher would choose, the next state should also be similar -- $T(s, \teacher(s)) \approx \generator(s, \agent(s))$.

For the generator part of the training (Lines~\ref{algo:line:step1_generator_begin}-\ref{algo:line:step1_generator_end}), \abbrev{} uses the agent samples collected from the environment to learn how to approximate the MDP's transitions using:
\begin{equation} \label{eq:lr_generator}
    \argmin_\generatorW \textstyle\!\sum_{\tau \in \Tau_\agent}\!\sum_{s \in \tau} \mid \transition(s, \agent(s)), \generator(s, \agent(s))) \mid.
\end{equation}
Due to the lack of knowledge about actions, it is impossible to approximate the transition function based on the teacher's observations alone. This part aims to enable the generator to learn how the MDP's transition behaves through transitions caused by the agent in the environment.

Therefore, by intercalating policy and generator training, \abbrev{} learns an approximate transition function that is leveraged to learn a policy similar to the teacher's.
As the agent approximates its policy to the teacher, so do its trajectories, leading the generator to fine-tune itself to similar transitions.
In other words, since the agent replicates trajectories close to the teacher and the generator only updates its weights via online play (i.e., by interacting with the environment), \abbrev{} ensures a positive correlation between the teacher and agent models.
Specifically, to minimise Eq.~\ref{eq:lr_policy}, the policy must predict an action that generates the same next state $\generator(s, \agent(s)) = \generator(s, \teacher(s))$, and to minimise Eq.~\ref{eq:lr_generator}, the generator has to correctly predict the next state based on the current state of the environment and the action predicted by the policy $\generator(s, \agent(s)) = \transition(s, \agent(s))$.
By not having any supervised losses, the policy must learn for itself which actions align more with the teacher's next expected state and better understand how different values affect the transition function from $\generator$.
Hence, the loss function does not introduce bias into the policy's action predictions, thereby addressing the supervised loss limitations of ILfO approaches.
Moreover, \abbrev{} circumvents the reliance on injective transition functions by learning from state transitions $P(s' \mid s, \agent(s))$ rather than direct state-action mappings $P(a \mid s)$.
This is because the gradients updating $\agentW$ originate from the generator itself (purple arrows in Fig.~\ref{fig:training_cycle}a, and the generator learns from state-action pairs.

Nevertheless, at the end of the reconstructive stage, the policy will have learned only from teacher transitions, and these may not be sufficient to determine how the agent should act in unseen cases.
To mitigate the absence of training data for these cases, we add an adversarial stage with online play to further align the policy with the teacher's and improve generalisation.

\subsection{Adversarial Stage} \label{sec:sub:adversarial_step}

The adversarial stage (Fig.~\ref{fig:training_cycle}b) trains both the policy, based on the agent samples collected from the environment $\Tau_\agent$, and the discriminator $\discriminator$ based on both the teacher samples $\Tau_\teacher$ and the agent's.
First, \abbrev{} freezes $\generator$ so that the model does not learn to ``fool'' the discriminator by diverging from the learned approximate transition function from the reconstructive stage.
In this stage, only the policy $\agent$ and discriminator $\discriminator$ are trained.
We freeze the generator $\generator$ to preserve the dynamics learned during the reconstruction phase.
This ensures that the policy alone is fine-tuned to better reproduce the teacher's behavioural patterns without altering the environment approximation.
If \abbrev{} were not to freeze its generator, given there are more optimisation functions binding $\generator$'s optimisation to the MDP's transition function, the generator would focus on learning how to force the discriminator to misclassify the states it generates, diverging from the learnt transition function $\generator = \transition' \nsubseteq \transition$.

Afterwards, \abbrev{} collects samples using $\agent$ in the environment (Line~\ref{algo:line:collect_agent_step2}) and trains both models by optimising a minmax framework:
\begin{equation} \label{eq:adversarial}
    \min_{\discriminatorW} \max_{\agentW}
    \mathbb{E}_{\tau \sim \Tau_\teacher^\Delta}\!\!\log(\discriminator(\tau)) \!+\!
    \mathbb{E}_{\tau \sim \Tau_\agent^\Delta}\!\!\log(1\!-\!\discriminator(\tau))
\end{equation}
where $\Tau_\teacher^\Delta$ is the sequence of the differences between subsequent states in teacher trajectories, and $\Tau_\agent^\Delta$ is the sequence of the differences between subsequent states generated by $\generator$.
Hence, the $i$th entry from $\Tau_\teacher^\Delta$ is $|s_i - \transition(s_i, \teacher(s_i))|$, and from $\Tau_\agent^\Delta$ is $|s_i - \generator(s_i, \agent(s_i))|$.
\abbrev uses the difference between states to avoid memorising the teacher's states since they are fixed during training.
Moreover, $\discriminator$ can be more applicable to unseen states by using the difference since the differences between states will hold.

It is important to note that the discriminator in \abbrev{} is a recurrent neural network.
Thus, it receives the differences between all pairs  of states in a trajectory and updates both the policy and its weights. 
Using the whole trajectory allows \abbrev{} to have a global view of the behaviour of the teacher and agent through the sequence of states and since the discriminator is recurrent, it can correct the policy when it acts optimally locally but not globally.
Nevertheless, the use of an adversarial stage poses some challenges: (i) it is susceptible to entering behaviour-seeking mode, and (ii) heavy updates to the policy can lead to \textit{catastrophic forgetting}, i.e., loss of previously learned information by a pre-trained model~\cite{Kirkpatrick2017forgetting}.
These challenges may arise because the adversarial stage encourages the agent to take actions that produce differences resembling those made by the teacher.
However, these differences may not be optimal across all possible trajectories. 
This can result in poorer trajectories, causing a compounding error between the policy and the discriminator, significantly altering $\agentW$.
As a consequence, the policy may lose its previously acquired knowledge.
To address these challenges, \abbrev{} uses only $10$ epochs in the adversarial stage ($\text{epochs}_a$), a drastically lower learning rate, and clips the gradients when updating $\agentW$.
By avoiding prolonged adversarial training, \abbrev{} manages to avoid behavioural-seeking mode, while keeping the acquired knowledge the policy learned from its reconstruction step.

\begin{table*}
    \footnotesize
    \centering
    \begin{tabular*}{\textwidth}{l@{\extracolsep{\fill}} c r r r r r}
        \toprule 
         Methods & Metrics & InvertedPendulum-v4 & Hopper-v4 & Ant-v4 & Swimmer-v4 & HalfCheetah-v4 \\ \midrule
        \multirow{1}{*}{Random ($\random$)} & AER & $5.7 \pm 3.26$ & $18.8985 \pm 18.4452$ & $-57.2408 \pm 97.069$ & $0.73 \pm 11.44$ & $-284.446 \pm 79.2153$ \\ 
        \multirow{1}{*}{Teacher ($\teacher$)} & AER & $1000 \pm 0$  & $3530.2857 \pm 12.2681$ & $5765.3842 \pm 621.1187$ & $355.4238 \pm 1.832$ & $9512.2995 \pm 538.5918$ \\ \midrule
        \multirow{2}{*}{\abbrev{}} & AER & $\mathbf{1000 \pm 0}$ & $\mathbf{3573.4266 \pm	8.5877}$ & $\mathbf{5904.2506 \pm 398.1026}$ & $\mathbf{361.2001 \pm 0.9960}$ & $\mathbf{9959.5745 \pm 589.3388}$ \\ 
                & $\performance$ & $\mathbf{1}$ & $\mathbf{1.0127}$ & $\mathbf{1.0238}$ & $\mathbf{1.0163}$ & $\mathbf{1.0457}$ \\\midrule 
        \multirow{2}{*}{CILO} & AER & $\mathbf{1000 \pm 0}$ & $3559.3540 \pm 10.3743$ & $5848.3754 \pm 458.6725$ & $356.8693 \pm 1.495$ & $9510.299 \pm 765.2461$ \\
                & $\performance$ & $\mathbf{1}$ & $1.0078$ & $1.0143$ & $1.0041$ & $0.9998$ \\  \midrule 
        \multirow{2}{*}{MAHALO} & AER & $\mathbf{1000 \pm 0}$ & $1033.9924 \pm 0.5959$ & $903.7594 \pm 37.1806$ & $87.4172 \pm 10.9691$ & $154.6271	\pm 4.0698$\\
                & $\performance$ & $\mathbf{1}$ & $0.2891$ & $0.1650$ & $0.0248$ & $0.0448$ \\ \midrule
        \multirow{2}{*}{OPOLO} & AER & $990.6254 \pm 13.8079$ & $3505.4686 \pm 40.9337$ & $5791.2084 \pm 786.6135$ & $352.2456 \pm 2.1989$ & $9442.2904 \pm 869.6332$ \\
                & $\performance$ & $0.9906$ & $0.9934$ & $1.0044$ & $0.9910$ & $0.9929$ \\ \midrule 
        \multirow{2}{*}{GAIfO} & AER & $870.3248 \pm 31.9845$ & $2586.4747 \pm 977.1989$ & $2656.604 \pm 963.7789$ & $360.5495 \pm 0.9515$ & $1988.5288 \pm 790.4208$ \\
                &  $\performance$ & $0.8696$ & $0.7315$ & $0.4661$ & $1.0145$  & $0.2320$ \\ \midrule 
        \multirow{2}{*}{BCO} & AER & $521.826 \pm 178.9$ & $1156.643 \pm 298.1488$ & $2324.0222 \pm 1573.7205$ & $347.1842 \pm 2.2823$ & $6776.0547 \pm 1755.5855$ \\
                & $\performance$ & $0.5191$ & $0.3241$ & $0.4090$ & $0.9768$ & $0.7207$ \\
        \bottomrule
    \end{tabular*}
    \caption{\abbrev and baselines AER and $\performance$ results for benchmarks as the average of $10,\!000$ seeds.}
    \label{tab:results}
\end{table*}

We note that \abbrev{} achieves state-of-the-art results even without the adversarial stage. However, this stage fine-tunes the policy further, by allowing it to perform online play in the environment. This leads to a slight increase in the reward and a decrease in the standard deviation between trajectories (as described in Sec.~\ref{sec:sub:role_of_each_step}).

\section{Experimental Results} \label{sec:experimental_results}

We compared \abbrev's results against the following five key related methods.
Behavioural Cloning from Observation (BCO)~\cite{torabi2018bco} and Generative adversarial Imitation from Observation (GAIfO)~\cite{torabi2018gaifo}, which are usually used as baselines, and three of the better ILfO methods: Continuous Imitation Learning from Observation (CILO)~\cite{gavenski2024explorative}, Modality-agnostic Adversarial Hypothesis Adaptation for Learning from Observations (MAHALO)~\cite{li2023mahalo}, and Off-Policy Imitation Learning from Observations (OPOLO)~\cite{zhu2020opolo}.
We do not compare to other baselines, such as DICE-based methods, such as SMODICE~\cite{ma2022smodice} and IQ-Learn~\cite{garg2021iq}, since MAHALO outperforms them in their experimentation.
We experimented with five commonly used environments: Ant, Half Cheetah, Hopper, Swimmer, and Inverted Pendulum.

We trained each method using the hyperparameters provided in the supplementary material and ran for $10,\!000$ randomly generated seeds not present in the training data. 
Each environment trajectory consists of $1,\!000$ steps or the number of steps before the agent falls.

\subsection{Implementation and Metrics} \label{sec:sub:implementation_and_metrics}

We used PyTorch to implement our agent and optimise the loss functions in Eq.~\ref{eq:lr_policy}-\ref{eq:adversarial} via Adam~\cite{kingma2014adam}.
We used the datasets provided by Imitation Datasets~\cite{gavenski2024ildatasets}, and Optuna~\cite{akiba2019optuna} for all hyperparameter searches.
It is important to note that we use Imitation Datasets for two reasons:
\begin{enumerate*}[label=(\roman*)]
    \item different from other datasets (e.g., D4RL~\cite{fu2020d4rl} and Minari~\cite{minari}), Imitation Datasets uses the newest version of these benchmarks (v$4$ instead of v$2$), allowing for easier reproducibility and longer support; and
    \item allows us to preserve the characteristics of the experiments across all evaluations by running them with the same seeds, guaranteeing replicability and having access to the teacher weights.
\end{enumerate*}
Yet, we note that Imitation Datasets only provides the teacher datasets and weights, and all benchmarks are the same in this work as others (e.g., MuJoCo).
We also provided all learning rates in the supplementary material with a link to the official implementation.

We evaluate all methods using the \textit{Average Episodic Reward} ($AER$) metric and  \textit{Performance} $\performance$.
$AER(\policy)$ is the average accumulated reward for a policy $\policy$ over $n$ trajectories with $t$ steps each:
$AER(\policy) = \frac{1}{n} \sum_{i=1}^{n}\sum_{j=1}^{t} \gamma^t r(s_{i,j}, \policy(s_{i, j})).$
$\performance_{\tau}(\policy)$ is the reward of the policy $\policy$ in a trajectory $\tau$, normalised between random and teacher policies, where $0$ corresponds to the reward of a random policy $\random$, and $1$ to the reward of the teacher policy $\teacher$:
$\performance(\policy) = \nicefrac{AER(\policy) - AER(\random)}{AER(\teacher) - AER(\random)}.$
Finally, we do not report accuracy since achieving high accuracy does not necessarily translate into a high reward for the agent.

To avoid any confusion, we note that although IL methods might achieve $\performance$ higher than one, the overall objective is not to be as close as possible to the teacher (e.g., $AER(\agent) \approx AER(\teacher)$), but to measure whether the agent is learning a policy that is relatively better than both policies.
Results in Tab.~\ref{tab:results} are the average and standard deviation in the five environments.
We do not report accuracy since achieving high accuracy does not necessarily translate into a high reward for the agent (seen in Sec.~\ref{sec:sub:ground_truth_approximation}).

\subsection{Results} \label{sec:sub:results}

Tab.~\ref{tab:results} shows the results from \abbrev and all baselines in $10,\!000$ different seeds.
Each seed is selected based on whether it appeared in the training data or during the online play for each method.
We select each seed based on its first state because it is impossible to guarantee that all methods will not predict actions to match its current trajectory to one of the teacher's.
Hence, we reset the environment and checked whether the first state appeared during the online play or in the entire training dataset; if it did not, we stored that seed for use during test time.
The first and second rows from Tab.~\ref{tab:results} show a random policy $\random$, and the teacher $\teacher$ used to generate each dataset in the same $10,\!000$ seeds.
Although the dataset's original $AER$ might differ from those shown in Tab.~\ref{tab:results} for the teacher, we believe comparing each agent's results under the same conditions is fairer whenever possible.
In particular, they should employ the same seeds and weights, which Imitation Datasets allows since it shares the teacher weights.
Therefore, the teacher row shows the $AER$ for the teacher in the same seeds as all other methods.

We observed that \abbrev was the only method that consistently achieved performance higher than $1$.
It was the only method to outperform its teacher in HalfCheetah, and by a large margin ($447.275$), whereas two of the other methods had difficulty even approximating the teacher's performance.
Moreover, \abbrev's standard deviation was lower than all other methods in all environments, only slightly higher than the teacher's in HalfCheetah and smaller or equal to the teacher's in all other environments.
We argue that this demonstrates that \abbrev can outperform the teacher and, more importantly, it deviates less from the desired behaviour.
\abbrev achieves an average performance of $1.0197$, the highest across all benchmarks, with CILO being the second with $1.0054$.
Note that $\performance$ needs to be analysed in tandem with coefficients of variation (CVs --- the ratio of the standard deviation to the mean average), because methods may be more sensitive to variations in trajectories, with lower CVs suggesting greater generalisability.
In addition to its superior performance in all environments, \abbrev also displays the lowest CV of all methods. 

CILO and OPOLO achieved the second and third best performances, $1.0054$ and $0.9945$, and their CVs were $3.32$ and $5.18$, respectively.
This was not surprising, given that CILO has been shown to outperform OPOLO in previous experiments~\cite{gavenski2024explorative}.
We observed that CILO lost performance when learning policies in environments with more strict clusters of actions (i.e., InvertedPendulum and HalfCheetah), confirming observations in~\cite{gavenski2024explorative}.
As for OPOLO, we noticed that performance was lost when the number of teacher samples was increased. 
We argue that this is due to its behaviour-seeking nature and its reliance on temporal information.
As the dataset grows in size, the dimensionality problem becomes more apparent.
Increasing the size also increases the chance of the same state appearing in different timesteps and being considered different (due to the temporal information). 
Yet, requiring fewer samples to learn is theoretically better if the model remains precise.
Unfortunately, this is not the case, as pointed out by \citeauthor{gavenski2024explorative}~\shortcite{gavenski2024explorative} when comparing their method and OPOLO with the same number of sampled trajectories.
To further test this,  we assessed \abbrev's ability to generalise in Sec.~\ref{sec:sub:sample_efficiency}, by evaluating its performance in trajectories with varying initial state distances and its sample efficiency.

GAIFO, BCO, and MAHALO had the worst performances, $0.6627$, $0.5899$, and $0.3047$, respectively.
This was expected for GAIFO and BCO, since they have more rudimentary mechanisms for handling trajectory variations, use action-based supervised losses, and adopt behaviour-seeking mode more often.
We did not expect their higher CV, despite their lower performance.
Usually, as ILfO methods fail to learn the teacher policy, their results also display lower CV (as indeed happened for MAHALO).
HalfCheetah was the only environment in which BCO outperformed GAIFO.
We interpret this as BC approaches generally being better at solving the environment task, as evidenced by CILO (a BC approach) outperforming OPOLO (which is not).
We argue that MAHALO's poor results are due to its reliance on external data with specific characteristics (see the Supplementary Material for a discussion, including an ablation study to validate our results).
\section{Theoretical Underpinning for Convergence}

Sec.~\ref{sec:experimental_results} presents the empirical evidence for the convergence of \abbrev over five benchmarks.
We now provide the theoretical explanation of convergence for \abbrev under the reconstruction and adversarial stages.

\subsection{\abbrev Setup}

We first start by considering:
\begin{enumerate*}[label=(\roman*)]
    \item $\agent$ is the policy model, computed by a neural network parametrised by $\agentW$. Given a state, it selects the action to be taken;
    \item the generative model $\generator$ is a neural network parametrised by $\generatorW$. It outputs the state reached by the execution of an action in a state;
    \item the discriminator model $\discriminator$ is a neural network parametrised by $\discriminatorW$. Given a trajectory, it outputs a value in the unit interval, estimating the likelihood that a trajectory belongs to the teacher.
\end{enumerate*}
Furthermore, let $\tau_\policy \sim \Tau_\policy$ be trajectories generated by a policy $\policy$ (either $\agent$ or $\teacher$), and $\Loss$ be the squared loss: $\Loss(s', \generator(s, \agent(s))) = \left\| s' - \generator(s, \agent(s)) \right\|^2$ (as stated in Sec.~\ref{sec:method}).

\subsection{Reconstruction Stage}

We begin by analysing the first stage from \abbrev: the reconstruction stage.
Lemma~\ref{lemma:local_convergence} provides the intuition behind local convergence.
\begin{lemma} \label{lemma:local_convergence}
    Let $\agent$ be the policy model, $\generator$ be the generative model, and $\Loss$ be the squared loss, which interleaves every $k$ iteration the parameter updates as follows:
    \begin{itemize}
        \item $\agentW_{k} \leftarrow \argmin_\agentW 
            \mathbb{E}_{\left(s, s'\right) \sim \tau_\teacher}
            \Loss(s', \mathcal{G}_{\phi_{k-1}}( \policy_{\theta_{k}}\!(s)))$; and 
        \item $\generatorW_k \leftarrow \argmin_\generatorW
            \mathbb{E}_{\left(s, s'\right) \sim \tau_{\policy_{\theta_{k}}}}
            \Loss(s', \mathcal{G}_{\phi_{k}}(\policy_{\theta_{k}}\!(s)))$.
    \end{itemize}
    We assume that $\agent$ and $\generator$ are continuously differentiable with Lipschitz-continuous gradients, and that learning rates are bounded so that parameter updates remain within compact sets.
\end{lemma}
\noindent
Then the parameters $\agentW_k$ and $\generatorW_k$, where $k$ is the $k$th iteration from the reconstruction stage, converge to a local stationary point ($\agentW^*$, $\generatorW^*$).
Thus, the alternating update scheme corresponds to block coordinate descent with differentiable and Lipschitz-continuous objectives.
Standard convergence results (e.g., \citet{Bertsekas1999GradientCI} and \citet{razaviyayn2013stochastic}) ensure convergence to a stationary point under these assumptions.

Lemma~\ref{lemma:local_convergence} ensures convergence of the joint optimisation for the first stage of both neural networks.
We now present Lemma~\ref{lemma:generator_consistency}, which justifies the role of $\generator$ individually within this setup.

\begin{lemma} \label{lemma:generator_consistency}
    Let $\policy_{\theta_k}$ be a fixed neural network trained up to the $k$ iteration ($\agentW_k$), and $\mathcal{G}_{\phi_{k-1}}$ be a neural network parametrised by $\generatorW_{k-1}$ up to the $k-1$ iteration using $\mathbb{E}_{(s, s') \sim \tau_\agent} \Loss(s', \mathcal{G}_{\phi_{k-1}}(s, \policy_{\theta_k}\!(s)))$ with sufficient model capacity and training data, the $k$th iteration of the generative update leads to $\mathcal{G}_{\phi_k}(s, \policy_{\theta_k}(s)) \rightarrow \mathbb{E}\left[s' \mid s, \policy_{\theta_k}\!(s)\right]$
\end{lemma}

\noindent
In other words, this setup is a supervised regression problem where the target is the next state under the transition dynamics $\generator(s, \agent(s)) \approx \transition(s, \agent(s))$.
We note that the reconstruction stage only updates $\generator$ weights from the environment since it is impossible to predict $\generator(s, \agent(s))$ with no knowledge of how $\transition(s, \agent(s))$ occurs.
Therefore, given universal approximation capacity, $\generator$ converges to the conditional expectation of the MDP's dynamics ($\transition$).
Moreover, as $\agent$ improves and generates trajectories closer to those in $\Tau_\teacher$, the distribution $\tau_\agent$ becomes increasingly aligned with $\tau_\teacher$, improving the generator's data quality.

Now, we present Lemma~\ref{lemma:agent_convergence}, which characterises the role of $\agent$ individually for the reconstruction stage.

\begin{lemma} \label{lemma:agent_convergence}
    Let $\mathcal{G}_{\phi_{k-1}}$ be the fixed generative neural network trained up to the $k-1$ iteration, and $\policy_{\theta_{k}}$ be the policy neural network parametrised by $\agentW_{k}$ up to the $k-1$ iteration. 
    Assume $\generator + \epsilon = \transition$, where $\epsilon$ is some error that inevitably decreases as $k$ increases on the support of $\agent$.
    Then, updating $\agentW_{k}$ via $\mathbb{E}_{(s, s') \sim \tau_\teacher} \Loss(s', \mathcal{G}_{\phi_{k-1}}(s, \policy_{\theta_{k}}(s)))$ leads to $\agent$ producing teacher-like transitions.
\end{lemma}

\noindent
If $\generator$ accurately approximates the environment's transition model, then matching $\generator(s, \agent(s))$ to the teacher-observed $s'$ encourages $\agent$ to choose actions that align with the teacher's behaviour.
As $\agent$ improves and produces more teacher-like state transitions, the generator receives better on-policy samples in future iterations, enhancing its approximation of the environment.
This recursive structure forms a positive feedback loop that enables consistent \textit{unsupervised} updates to $\agentW$.
It is important to note that \abbrev learns the teacher's actions unsupervisedly in an unsupervised way via this process, and $\agent$ only learns from teacher transitions via $\generator(s, \agent(s))$.
Therefore, no direct action signal is backpropagated to the neural network at any time.

Lemmas~\ref{lemma:generator_consistency} and~\ref{lemma:agent_convergence} show that improvements to the generator reinforce the policy's ability to mimic teacher behaviour and vice-versa.
We can now explore the convergence of this coupled process.

\begin{lemma} \label{lemma:recursive_convergence}
    Let ($\agentW_k$, $\generatorW_k$) be the parameters at iteration $k$ of the alternating optimisation previously defined.
    Under boundedness of updates and the assumptions of Lemmas~\ref{lemma:local_convergence}-\ref{lemma:agent_convergence}, ($\agentW_k$, $\generatorW_k$) converges to ($\agentW^*$, $\generatorW^*$) such that $\tau_\agent + \epsilon = \tau_\teacher$ where $\epsilon$ is an acceptable error assuming both $\tau_\agent$ and $\tau_\agent$ start in the same state $s$.
\end{lemma}
\noindent
In other words, Lemma~\ref{lemma:recursive_convergence} follows from the two-timescale stochastic approximation theory~\cite{Borkar2008StochasticAA}.
The generator adapts quickly to the current policy, and the policy updates using a fixed generator, where the joint process converges to a consistent fixed point.

\subsection{Adversarial Stage}

After the reconstruction stage, an adversarial stage refines the policy $\agent$ using a discriminator $\discriminator$.
Lemma~\ref{lemma:optimal_discriminator} explores the discriminator optimisation under this stage.

\begin{lemma} \label{lemma:optimal_discriminator}
    Let $\tau$ denote a trajectory, which is a finite sequence of states $\tau = \left(s_0, s_1, \ldots, s_t \right)$, $p_\policy(\tau)$ denote the probability densities over trajectories by the teacher policy $\teacher$ and the agent's $\agent$, and the discriminator $\discriminator$ be trained under the loss function: 
    \begin{equation*}
        \argmax_\discriminatorW \mathbb{E}_{\tau_\teacher\!\sim\!\Tau_\teacher}\!\log\!\discriminator(\tau_\teacher) + \mathbb{E}_{\tau_\agent\!\sim\!\Tau_\agent}\!\log(1\!-\!\discriminator(\tau_\agent)).
    \end{equation*}
    Then, for a fixed policy $\agent$ and $\teacher$, the optimal $\discriminator$ is: 
    \begin{equation*}
        \discriminator^*(\tau) = \frac{p_\teacher(\tau)}{p_\teacher(\tau) + p_\agent(\tau)}.
    \end{equation*}
    Thus, minimising the result value function for $\agent$ corresponds to minimising the Jensen-Shannon divergence, $JSD(\teacher \mid\mid \agent)$, between the teacher and agent trajectory distributions.
\end{lemma}
\noindent
Lemma~\ref{lemma:optimal_discriminator} follows standard results in generative adversarial learning~\cite{goodfellow2014generative}, where the optimal discriminator for two distributions is known to be $\discriminator^*$.
The generator (in our work, $\generator(s, \agent(s))$) trains to match the teacher by minimising the Jensen-Shannon divergence.

While Lemma~\ref{lemma:optimal_discriminator} establishes the objective of the adversarial stage, it does not guarantee that the policy remains stable after reconstruction.
Therefore, Lemma~\ref{lemma:fine_tune} ensures that under mild constraints, fine-tuning does not lead to catastrophic forgetting or divergence.

\begin{lemma} \label{lemma:fine_tune}
    Assume that during adversarial training, the generator $\generator$ remains fixed after reconstruction, the discriminator $\discriminator$ is near-optimal at each step, and the policy $\agent$ is updated using small, clipped gradients.
    Then the adversarial stage preserves the behavioural properties of the reconstructed policy and improves generalisation without destabilising earlier learning.    
\end{lemma}

\noindent
In other words, policy updates are clipped and learning rates are kept small to avoid destabilising the pre-trained policy.
Therefore, the fixed generator ensures that the transition model used to compute gradient signals remains stable.
With a near-optimal discriminator, the gradients provided to $\agent$ reliably reflect the discrepancy between agent and expert trajectories.
Gradient clipping and a reduced learning rate guarantee that updates to $\agent$ are conservative, preventing large deviations from the pre-trained policy.
Under these conditions, the adversarial stage acts as a regularised adjustment rather than a full retraining, mitigating the risk of catastrophic forgetting~\cite{goodfellow2013empirical} while refining behavioural fidelity.

Finally, we note that these results only assume differentiability, Lipschitz continuity, and bounded updates.
No convexity assumptions are made.
Convergence should be interpreted as local convergence to stationary points, which are typical in deep learning settings.
\section{Discussion} \label{sec:discussion}

In this section, we consider some key aspects of \abbrev's behaviour:
\begin{enumerate*}[label=(\roman*)]
    \item the interaction between generator and policy in the first stage;
    \item how the adversarial step improves \abbrev overall results; and
    \item how sample efficient \abbrev is.
\end{enumerate*}

\subsection{Teacher Action Approximation} 
\label{sec:sub:ground_truth_approximation} 

It is vital for \abbrev{} to estimate action effects using solely the loss in Eq.~\ref{eq:lr_policy} closely via the process described in Sec.~\ref{sec:sub:reconstruction_step}.
If the signal from the generator is not strong enough and the policy is unable to approximate its predictions, or the generator is unable to learn a transition function close enough to the MDP's, the adversarial stage will not work, and \abbrev{} will not achieve the desired results.

\begin{figure}[b!tp]
    \centering
    \includegraphics[width=\linewidth]{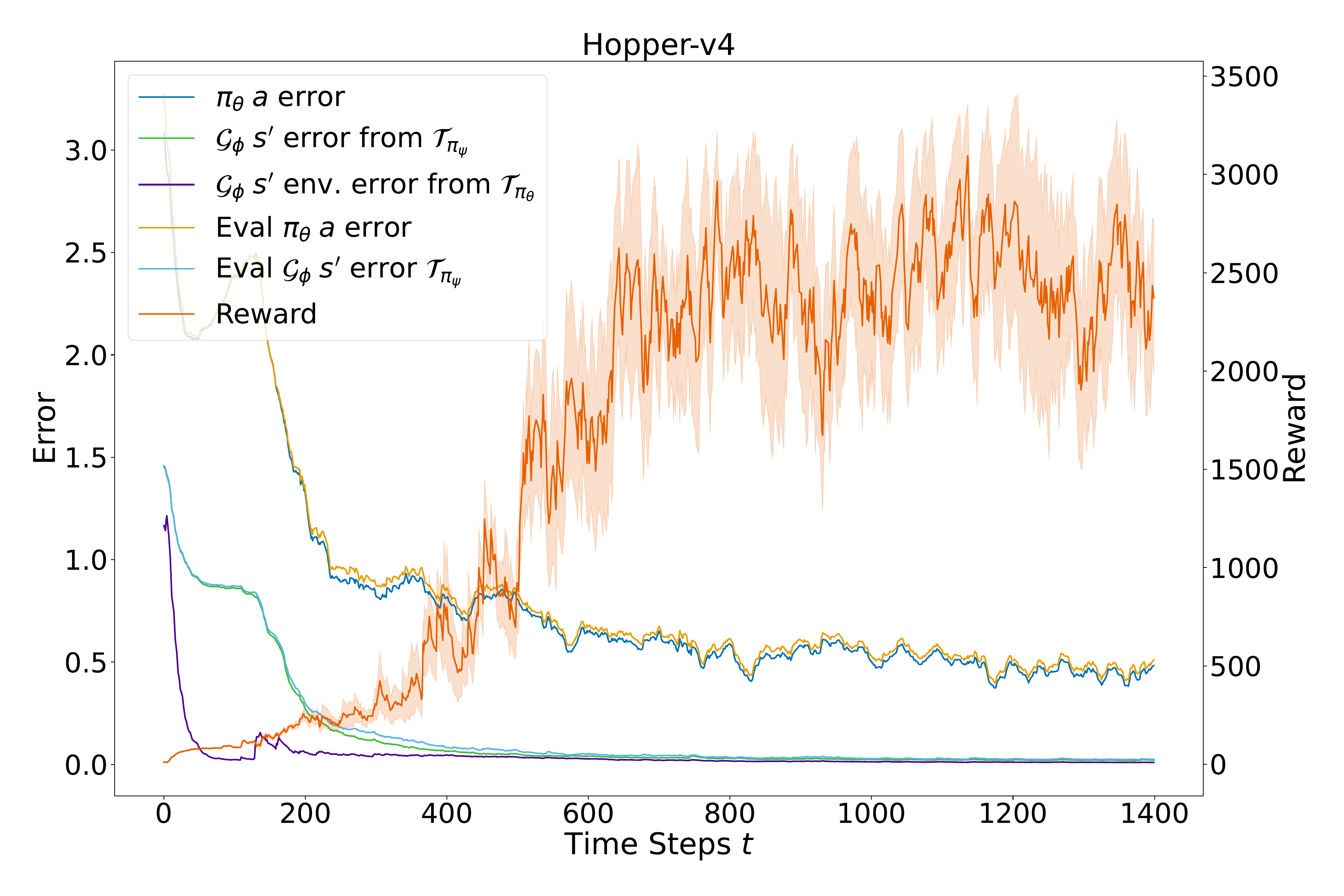}
    \vspace{-6pt}
    \caption{Training information for \abbrev in Hopper-v$4$.}
    \Description[Training curves for reconstruction stage]{
        Training metrics over time showing policy error, generator errors (training and evaluation), and agent reward for Hopper-v4. The generator error decreases and stabilises early, while the policy error shows a correlated decrease before stagnating. Reward increases as errors decrease, with shaded confidence intervals showing variability across runs.
    }
    \label{fig:ground_truth_approximation}
\end{figure}

In Fig.~\ref{fig:ground_truth_approximation}, we show the training behaviour for the Hopper-v$4$ environment during the reconstruction stage with a smoothing function for better visibility.
It displays the error from $\agent$ according to the teacher's dataset, the generative error from $\generator$ according to the teacher's dataset (split into training and evaluation) and the agent's collected samples, and the reward accumulated for the agent during each training step.
As expected, the generator learns a transition function that approximates the actual MDP transition function quite early in the training phase (purple line).
However, for the generator to see transitions closer to the teacher's, the policy has to improve its error, and we notice a direct correlation between the errors of $\agent$ and $\generator$.
We also observe that as the errors of both models decrease, the rewards increase, which shows that Eq.~\ref{eq:lr_policy} and~\ref{eq:lr_generator} yield the expected outcome.
Yet, we note that the policy error stagnates after a certain number of epochs, potentially indicating that the gradient signal from the generator is no longer enough to improve the policy's predictions further.
This behaviour is expected since the generator is not optimised to improve the policy's predictions themselves, but rather to learn a transition function that can approximate the teacher's and the MDP's.
Thus, the adversarial stage is crucial for further improving the policy's predictions and for stabilising the model's behaviour.
Finally, we believe that this stage could be further improved by using a curriculum learning approach. 
In this approach, the $\generator$ would receive more samples it predicted poorly, and the policy would receive more meaningful weight updates.

\subsection{Role of Each Stage} \label{sec:sub:role_of_each_step}

Tab.~\ref{tab:role_for_each_step} shows \abbrev and teacher results in the Half-Cheetah-v$4$ environment, the only in which no baseline achieved teacher performance.
This environment clearly displays the role of \abbrev's two stages.
The reconstruction stage achieves performance above $1$, with over $307$ more reward points than the teacher. The $AER$ of $9819.6491$ here with standard deviation of $626.0181$ are already superior to all other methods' (see Tab.~\ref{tab:results}).
However, the standard deviation is significantly higher than the teacher's (by $12.59\%$).
This is not ideal because we want a model that can perform at least as well as the teacher in all different environment settings.
\abbrev's adversarial stage improves $\performance$ and $AER$ further, whilst reducing its standard deviation by $5.85\%$.

\begin{table}[ht!]
    \scriptsize
    \centering
    \begin{tabular*}{\columnwidth}{l@{\extracolsep{\fill}}rrr}
        \toprule
        Step & \multicolumn{1}{c}{$AER$ ($\uparrow$)} & \multicolumn{1}{c}{CV ($\downarrow$)} & \multicolumn{1}{c}{$\performance$ ($\uparrow$)} \\ \midrule
         Teacher ($\teacher$) & $9512.2995 \pm 538.5918$ & \textbf{5.66}\% & 1.0000 \\ \midrule
         Reconstruction (Sec.~\ref{sec:sub:reconstruction_step}) & $9819.6491 \pm 626.0181$ & 6.38\% & 1.0314 \\
         Adversarial (Sec.~\ref{sec:sub:adversarial_step}) & $\mathbf{9959.5745 \pm 589.3388}$ & 5.92\% & \textbf{1.0457} \\
         \bottomrule
    \end{tabular*}
    \caption{HalfCheetah-v$4$ environment results.}
    \label{tab:role_for_each_step}
\end{table}

As mentioned in Sec.~\ref{sec:sub:adversarial_step}, the adversarial stage (Algorithm~\ref{algo:method}, Line~\ref{algo:line:step2_begin}--\ref{algo:line:step2_end}) might lead to behavioural-seeking mode and a catastrophic forgetting of the policy.
Therefore, it is preferable to run this stage for only a few epochs ($\text{epochs}_a \leq 10$) to reduce the policy's standard deviation through online play and slightly improve the agent's performance. 
Given that it only runs for a few epochs, it offers an excellent trade-off between time and policy improvement.

\subsection{Sample Efficiency} \label{sec:sub:sample_efficiency}

\begin{figure}[t!bp]
    \centering
    \includegraphics[width=\linewidth]{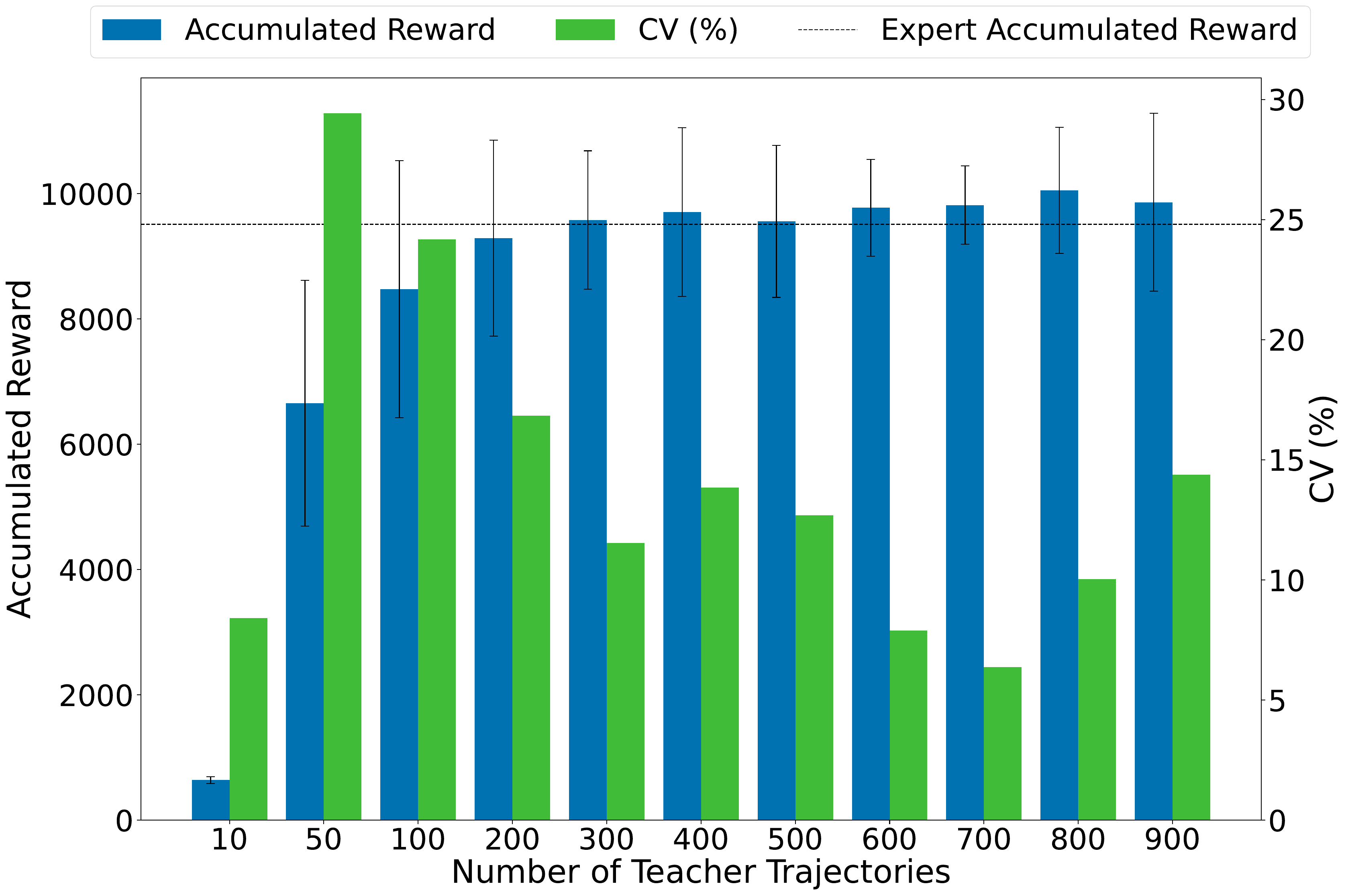}
    \caption{\abbrev's CV and $\mathbf{AER}$ for the HalfCheetah-v$4$.}
    \Description[Performance vs generalization trade-off across dataset sizes]{
        Bar chart showing accumulated reward and coefficient of variation (CV) for varying numbers of teacher trajectories in HalfCheetah-v4. Blue bars show reward increasing with dataset size, surpassing teacher performance (dashed line) around 300 trajectories. Green bars show CV initially high, decreasing until 800 trajectories where overfitting begins. Results demonstrate a trade-off between performance and generalisation, with optimal dataset size balancing maximum reward and minimum CV.
    }
    \label{fig:sample_efficiency}
\end{figure}

If a large number of teacher samples is required for \abbrev to achieve teacher performance, using it in environments where sample collection is challenging or costly may be impractical. 
However, performance decay with an increase in the number of samples indicates an issue with the behaviour-seeking aspect.
To see how \abbrev handles varying amounts of samples, we experimented with $10$ to $900$ trajectories.
Fig.~\ref{fig:sample_efficiency} shows the reconstruction stage results for all experiments in the HalfCheetah-v$4$ environment.

Although \abbrev does not achieve teacher performance on $10$ trajectories (unlike OPOLO and CILO), its AER and $\performance$ improve when increasing training data.
It eventually surpasses teacher performance with $300$ trajectories, albeit with a standard deviation higher than all other baselines.
As we increase the number of trajectories further, we observe a diminishing return from \abbrev's AER.
This would be a significant drawback were the method not to become more generalisable.
\abbrev's CV keeps decreasing up to around $800$ trajectories, when the method seems to overfit with a simultaneous increase in variation and AER.
We conclude that the ideal number of samples is the one that maximises AER while minimising CV (i.e., maximising generalisability).
Although other methods achieve performance comparable to their teacher on smaller datasets, \abbrev is the only one that consistently improves generalisation as the dataset size increases.

\section{Related Work} \label{sec:related_work}

The simplest form of imitation learning is BC~\cite{pomerleau1988alvinn}, which addresses the IL problem as a supervised learning problem.
It uses teacher samples of the form  $(s, a, s')$, where $s'=T(s,a)$, to learn how to approximate the teacher's policy.
However, supervised approaches are costly for more complex scenarios, requiring more samples about the actions' effects in the environment. 
For example, solving Atari games requires approximately $100$ times more samples than control tasks, such as CartPole.
Generative Adversarial Imitation Learning (GAIL)~\cite{ho2016generative} matches the state-action frequencies from the agent to those seen in the training data, creating a policy with action distributions closer to the teacher.
GAIL leverages the data structure more efficiently, reducing its dependency on large quantities of data and making it more suitable for complex environments.
It uses adversarial training to discriminate state-action pairs from the agent or the teacher while minimising the difference between them.

Recent self-supervised approaches~\cite{torabi2018bco,monteiro2023self} that learn from observations use teacher's transitions $(s_\teacher, s'_\teacher)$ and leverage random transitions $(s_\random, s'_\random)$ to learn the IDM of the environment, and afterwards, generate self-supervised labels for the teacher's trajectories.
Generative Adversarial Imitation Learning from Observation (GAIfO)~\cite{torabi2018gaifo} is an ILfO method based on  GAIL.
The discriminator is trained using state transitions, and the policy is optimised using an inverse reinforcement learning strategy.
Imitating Latent Policies from Observation (ILPO)~\cite{edwards2019ilpo} differs from such work by trying to estimate the probability of a latent action given a state.
Within some environment steps, it remaps latent actions.
More recently, Off-Policy Learning from Observations (OPOLO)~\cite{zhu2020opolo} used a dual-form of the expectation function and an adversarial structure to achieve off-policy ILfO.
Modality-agnostic Adversarial Hypothesis Adaptation for Learning from Observations (MAHALO)~\cite{li2023mahalo} unifies ILfO with offline reinforcement learning by learning pessimistic value estimates through adversarially trained reward and critic functions.
It accommodates partially labelled datasets and generalises across several ILfO settings.
Continuous Imitation Learning from Observation (CILO)~\cite{gavenski2024explorative} used the same IDM structure of earlier methods but introduced an exploration mechanism to further improve its IDM's predictions. 
Path signatures were used as non-parametric encodings to identify trajectories that behave similarly to the teacher efficiently.

\section{Conclusion} \label{sec:conclusion}

In this work, we proposed \method (\abbrev), an unsupervised ILfO method addressing critical limitations of existing approaches. 
Firstly, \abbrev eliminates reliance on action-based supervised optimisation, reducing the risk of overfitting. 
Secondly, it mitigates behaviour-seeking that often lead to suboptimal performance. 
Lastly, \abbrev provides generalisation by more effectively handling environments where transitions can result from multiple actions. 
It learns to approximate the teacher's policy and refine it through adversarial learning. 
A two-stage training process ensures the agent captures accurate state transitions and aligns its policy with the teacher's through online interactions. 
By leveraging transitions rather than direct state-action mappings, \abbrev avoids common ILfO pitfalls and achieves superior results across diverse initial states.

Our experiments highlight the effectiveness of \abbrev, consistently surpassing the teacher in five benchmark environments.
Unlike traditional methods, \abbrev achieves state-of-the-art results without sacrificing robustness, making it suitable for real-world scenarios where action labels are unavailable or costly. 
It demonstrates strong generalisation, maintaining low trajectory variance as evidenced by its coefficient of variation ($2.64\%$).
Beyond imitation learning, \abbrev offers a general framework for robust policy learning from indirect supervision.
Its proxy-based learning approach is applicable to offline RL, sim-to-real transfer, and causal representation learning in domains such as healthcare and economics. Moreover, the refinement stage resembles curriculum learning and can be adapted to enhance sample efficiency and robustness in safety-critical systems.

Future work will explore \abbrev's application to more complex and dynamic environments, further enhancing sample efficiency through advanced exploration strategies and expanding its applicability as a foundation for observational learning in autonomous systems.



\begin{acks}
This work was supported by UK Research and Innovation [grant number EP/S023356/1], in the UKRI Centre for Doctoral Training in Safe and Trusted Artificial Intelligence  (\url{www.safeandtrustedai.org}) and made possible via King's Computational Research, Engineering and Technology Environment (CREATE)~\cite{create}.
\end{acks}



\bibliographystyle{ACM-Reference-Format}
\bibliography{references}


\newpage
\appendix
\begin{table*}[ht]
    \scriptsize
   \centering
   \begin{tabular*}{\textwidth}{@{\extracolsep{\fill}}lc@{\extracolsep{\fill}}lc@{\extracolsep{\fill}}lc@{\extracolsep{\fill}}}
        \toprule 
        \multicolumn{2}{c}{$\agent$}                    & \multicolumn{2}{c}{$\generator$} & \multicolumn{2}{c}{$\discriminator$} \\
        \cmidrule{1-2} \cmidrule{3-4} \cmidrule{5-6}
        Layer Name          & Input $\times$ Output     & Layer Name        & Input $\times$ Output     & Layer Name           & Input $\times$ Output \\
        Input               & $\mid d \mid \times 400$  & Input               & $\mid d \mid \times 400$  & LSTM               & $\mid \beta \mid \times\; 64$ \\
        Activation (Tanh)   & -                         & Activation (Tanh)   & -                         & Fully Connected  1 & $64 \times 1024$   \\
        Fully Connected 1   & $400 \times 400$          & Fully Connected 1   & $400 \times 400$          & Leaky ReLU         & -                  \\
        Activation (Tanh)   & -                         & Activation (Tanh)   & -                         & Dropout            & 0.5\%              \\
        Fully Connected 2   & $400 \times 400$          & Fully Connected 2   & $400 \times 400$          & Fully Connected  2 & $1024 \times 1024$ \\
        Activation (Tanh)   & -                         & Activation (Tanh)   & -                         & Leaky ReLU         & -                  \\
        Fully Connected 3   & $400 \times 400$          & Fully Connected 3   & $400 \times 400$          & Dropout            & 0.5\%              \\
        Activation (Tanh)   & -                         & Activation (Tanh)   & -                         & Fully Connected  3 & $1024 \times 2$    \\
        Output              & $400\; \times \mid a \mid$  & Output              & $400 \times \mid a \mid$  & \\
        \bottomrule
   \end{tabular*}
   \caption{Parameters for \method{}.}
   \label{tab:params}
\end{table*}

\section{Environments}
In this work, we experiment with different environments.
We now briefly describe each environment.
We used Imitation Datasets~\cite{gavenski2024ildatasets} to gather teacher samples and its weights loaded from HuggingFace.\footnote{https://huggingface.co/}
All results are displayed in Tab.~\ref{tab:results} in Sec.~\ref{sec:sub:results} of the main work.
We used $700$ teacher trajectories for training and the remaining $300$ for validation.
All environment seeds used in this work are available at: \anonimity

\subsection{A note about the teacher samples}

Although all teachers are loaded from Imitation Datasets~\cite{gavenski2024ildatasets} share similar results, the behaviour of each teacher varies drastically.
One could argue that humans also deviate from each trajectory. However, using episodes with more human-like trajectories (where the robot movements were less hectic) yielded better results for all imitation learning approaches.
By being more consistent with their movements, teachers allow policies to receive more varied samples and generalise better.
All samples used in this work are available in \citeauthor{gavenski2024ildatasets}'s~\shortcite{gavenski2024ildatasets} repository.

\subsection{Ant-v4} 
Ant-v$4$ consists of an ant-like robot made out of a torso with four legs attached to it, with each leg having two joints~\cite{schulman2015high}.
The goal of this environment is to coordinate the four-legged robot to move to the right of the screen by applying force on the eight joints.
Ant-v$4$ requires eight actions per step, each limited to continuous values between $[-1, 1] \in \mathbb{R}$.
Its observation space consists of 27 attributes for the $x$, $y$ and $Z$ axis of the $3$D robot.
Imitation Datasets use TD$3$ weights as the teacher.

\subsection{Swimmer-v4} 
Swimmer-v$4$ consists of a robot with $s$ segments ($s \geq 3$) and $j = s - 1$ joints~\cite{coulom2002reinforcement}.
Following previous work~\cite{zhu2020opolo,gavenski2024explorative}, we use the default setting $2 = 3$ and $j = 2$.
The agent applies force to the snake-like robot's joints, and each action can range $[-1, 1] \in \mathbb{R}$.
A state is encoded by an $8$-dimensional vector representing all segments' angle, velocity and angular velocity.
Imitation Datasets use TD$3$ weights as their teacher.
Like Ant-v$4$, the agent's goal in this environment is to move as fast as possible to the right.
However, now, it does so by applying torque to the joints and using the fluid's friction.

\subsection{Hopper-v4} 
Hopper-v$4$ is a one-legged and two-dimensional robot with four main parts connected by three joints: (i) a torso at the top, (ii) a thigh in the middle, (iii) a leg at the bottom, and (iv) a single foot facing right~\cite{erez2011infinite}.
The agent's goal again is to make the robot move to the right (forward) by hoping.
Imitation Datasets use TD$3$ weights as the teacher, and each action is limited between $[-1, 1] \in \mathbb{R}$.

\subsection{HalfCheetah-v4} 
HalfCHeetah-v$4$ has a two-dimensional cheetah-like robot with two ``paws''~\cite{wawrzynski2009cat}.
The robot contains $9$ segments and $8$ joints.
Its actions are a vector of $6$ dimensions, consisting of the torque applied to the joints to make the cheetah move.
All states consist of the robot's position and angles, velocities and angular velocities for its joints and segments.
The agent's goal is to run to the right of the screen (forward) as fast as possible.
A positive reward is allocated based on the distance traversed, and a negative reward is awarded when moving to the left of the screen.
Imitation Datasets use TD$3$ weights as the teacher.
The action space is limited between $[-1, 1] \in \mathbb{R}$.


\subsection{InvertedPendulum-v4}
This environment is based on the CartPole environment from~\citeauthor{sutton2018reinforcement}~\cite{sutton2018reinforcement}.
It involves a cart that can move linearly, with a pole attached.
The agent can push the cart left or right to balance the pole by applying forces on the cart.
The goal of the environment is to prevent the pole from reaching a particular angle on either side.
The continuous action space varies between $-3$ and $3$, the only one within the five environments outside the $-1$ to $1$ limit.
Its observation space consists of $4$ different attributes.
We use Stable Baselines~$3$'s PPO weights.
The expert sample size is $10$ trajectories, which consist of $10,000$ states (with their $4$ attributes) and actions (with a single action value per step).
The invertedPendulum-v$4$ environment is the only one that has an expert with the environment's maximum reward.
Therefore, achieving $\mathcal{P}$ higher than $1$ is impossible.

\section{Hyperparameters}
For training, we used an NVidia A$100$ GPU and PyTorch.
Although we used this GPU, such hardware is not strictly required since \abbrev uses less memory to train.
Tab.~\ref{tab:lr} shows the different learning rates used in the main work for each environment, where $\Loss_R$ is the learning rate used during the reconstruction step to train $\agent$ and $\generator$, $\Loss_{\agent}$ is the learning rate for updating $\agentW$ during the adversarial step, and $\Loss_{\discriminator}$ is the learning rate for updating $\discriminatorW$ during the adversarial step.

\begin{table}[ht]
    \scriptsize
    \centering
    \begin{tabular*}{\columnwidth}{c@{\extracolsep{\fill}} rrrrr}
    \toprule
    Learning Rate & \multicolumn{1}{c}{Swimmer} & \multicolumn{1}{c}{Hopper} & \multicolumn{1}{c}{Ant} & \multicolumn{1}{c}{Swimmer} & \multicolumn{1}{c}{HalfCheetah} \\ \midrule
    $\Loss_R$ & 2.01E-04 & 2.34E-05 & 2.01E-04 & 4.70E-05 & 4.96E-04 \\
    $\Loss_{\agent}$ & 1.18E-05 & 6.07E-06 & 1.18E-05 & 1.47E-05 & 1.80E-05 \\
    $\Loss_{\discriminator}$ & 2.41E-03  & 9.97E-04 & 2.41E-03 & 2.67E-03 & 4.33E-03 \\
    \bottomrule
    \end{tabular*}
    \caption{Learning rates used for each environment.}
    \label{tab:lr}
\end{table}

\section{Network Topology}
Tab.~\ref{tab:params} shows the network topology for all \abbrev neural networks.
The $\generator$ and $\agent$ are Multi-Layer Perceptron (MLP) networks with $3$ fully connected layers, each with $400$ neurons, except for the last layer whose size is the same as the number of environment state and action dimensions, respectively.
The $\discriminator$ is a Long Short-Term Memory (LSTM)~\cite{graves2012long} network with $64$ hidden state neurons and $2$ layers, followed by a classification head of $2$ fully connected layers, each with $1024$ neurons and dropout.
We use a Tanh activation function for $\generator$ and $\agent$ since the environment's actions are between $[-1, 1]$.
For $\discriminator$, we use Leaky ReLU, the standard for classification tasks.
During our experimentations, we discovered that when trying to use Self-Attention layers, as used in \cite{gavenski2024explorative}'s work~\shortcite{gavenski2024explorative}, it did not work.
The neural networks would drastically overfit when these mechanisms were used within the policy and would not learn when used in the generator and policy.
An official implementation of \abbrev is available at: \anonimity

\subsection{Would unfreezing $\generator$ in Stage 2 work?} \label{sec:sub:unfreezing_generator}

Sec.~\ref{sec:sub:adversarial_step} mentioned that to improve the agent consistency \abbrev freezes $\generator$ during the adversarial stage.
To understand why this is necessary, we perform an experiment that does not freeze $\generator$ during the adversarial stage for the Swimmer-v$4$ environment.
We note that for this experiment, we allow the adversarial stage to train for longer since the dynamics between models change.
With the generator updating its weights, it is theoretically possible to learn a better policy with more time than its counterpart.
Yet, we do not expect that to happen, as mentioned previously.

Tab.~\ref{tab:unfreezing} shows \abbrev results when freezing and unfreezing $\generator$.
As expected, the version where $\generator$ was frozen outperformed the one where it was not.
We observe that for the Swimmer-v$4$ environment, the version where $\generator$ was not frozen also outperformed its teacher.
However, unfreezing $\generator$ caused a significant increase in \abbrev variation ($0.29$ points higher) since its AER decreases.
We interpret this as the generator model learning how to ``fool'' the discriminator and unlearning its initial transition function $\transition' \subseteq \transition$.
We also observe that by unfreezing $\generator$, \abbrev achieves CV higher than the teacher, which is undesirable.
Lastly, we note that the results for the unfrozen $\generator$ used the best policy's weights before it collapsed due to catastrophic forgetting.

\begin{table}[ht!]
    \scriptsize
    \centering
    \begin{tabular*}{\columnwidth}{l@{\extracolsep{\fill}}rrr}
        \toprule
        $\generator$ & \multicolumn{1}{c}{AER ($\uparrow$)} & \multicolumn{1}{c}{CV ($\downarrow$)} & \multicolumn{1}{c}{$\performance$ ($\uparrow$)} \\ \midrule
        Teacher ($\teacher$) & $355.4238 \pm 1.832$ & 0.52\% & 1 \\ \midrule
        Unfrozen & $356.1365 \pm 2.0129$ & 0.57\% & 1.0020 \\
        Frozen & $\mathbf{361.2001 \pm 0.9960}$ & \textbf{0.28\%} & \textbf{1.0163} \\
        \bottomrule
    \end{tabular*}
    \caption{Results for the Swimmer-v$4$ when unfreezing $\generator$.}
    \label{tab:unfreezing}
\end{table}

\section{On The Use Of External Data}

\abbrev does not rely on any external data beyond the teacher's observations.
It collects all additional samples used during training through the agent's interactions with the environment.
In contrast, ILfO methods commonly use external data to bridge the gap between their initial random policy and the teacher's policy.
They do so by collecting samples from a random policy to learn the transition dynamics and to utilise teacher samples in a self-supervised manner.
MAHALO~\cite{li2023mahalo} slightly differentiates from this setting.
Although MAHALO is a more recent imitation learning method, its performance in our evaluation is notably lower.
We attribute this to three key differences:
\begin{enumerate*}[label=(\roman*)]
    \item differences in problem formulation between MAHALO, \abbrev, and other ILfO methods,
    \item the data and hyperparameters used, and
    \item our interpretation of the discrepancy between MAHALO's reported performance and that observed in our setup.
\end{enumerate*}
In this section, we examine these differences to understand the role that external data plays in ILfO methods and how \abbrev differentiates itself from other ILfO methods.

\subsection{Problem Formulation}

The core difference between \abbrev, MAHALO, and other ILfO methods lies in (i) whether they require samples outside the teacher's ones, and (ii) how they utilise them.
\abbrev operates in a fully unsupervised manner, relying exclusively on the teacher trajectories and the agent's own interactions.
It does not leverage any form of labelled data, including action or reward information, nor does it rely on externally sourced trajectories to learn environment dynamics.
In contrast, MAHALO depends on an external offline RL dataset (e.g., D4RL~\cite{fu2020d4rl}) that contains labelled transitions, including both actions and rewards.
It utilises this dataset to train an inverse dynamics model, which is then applied to infer actions based on the teacher’s state-only demonstrations. 
This positions MAHALO closer to methods such as BCO~\cite{torabi2018bco}, GAIfO~\cite{torabi2018gaifo}, CILO~\cite{gavenski2024explorative}, which similarly infer actions or dynamics from external sources before applying them to the teacher data.
A key distinction introduced by this formulation is that MAHALO's reliance on offline datasets is populated with suboptimal but structured policies.
Unlike purely random trajectories, these suboptimal policies offer meaningful coverage of the environment and make the learning task considerably easier by implicitly providing informative intermediate behaviours.
It is this availability of suboptimal policy data -- not simply the act of bridging the distributional gap between random and teacher samples -- that underpins MAHALO's performance advantages.
This departs from the strict observational constraints upheld by \abbrev and other ILfO methods considered in our main work.

\subsection{Data and Hyperparameters}

With this clear distinction in mind, we now outline the hyperparameter selection process employed in our evaluation and the data used for the offline and online training portions.
In their original work, \citeauthor{li2023mahalo} state that the confidence threshold $\beta$, which determines whether inferred actions are accepted based on their likelihood, is tuned individually for each environment.
Unfortunately, no specific $\beta$ values are provided for the overlapping environments in our study (\textit{Hopper} and \textit{HalfCheetah}). 
Furthermore, our evaluation includes additional environments not present in their original setup (\textit{InvertedPendulum}, \textit{Ant}, and \textit{Swimmer}).
To ensure a fair and robust comparison, we train over all $\beta$ values reported in their supplementary material $\{0.1; 1; 10; 100; 1,000\}$.
For each environment, MAHALO is trained using the same $700$ teacher episodes employed across all baselines (which is approximately $300$ episodes short of their original work), and exclusively with random samples (i.e., no suboptimal policies are used from the offline data).
Although \abbrev does not use any external data, we maintain the use of random policy information to have a fairer comparison with other ILfO methods.
The random dataset contains $10,000$ episodes containing a tuple of $\tuple{s, a, s', r}$, and whether the state was a terminal state or not.
We use the same random dataset for all methods to maintain consistency, and the same teacher datasets, which are the v$4$ version for MuJoCo (present in IL-Datasets~\cite{gavenski2024ildatasets}) instead of the v$2$ used in MAHALO's original work.
We conduct evaluation using separate seeds, not seen during training and not used for testing.
The $\beta$ value yielding the best performance (measured by $AER$) is selected.
Each experiment retains the original $\alpha/\beta$ ratio of $100,000$ as this hyperparameter is held constant across all environments in the original work.
We report the selected $\beta$ values, along with their $AER$ and $\performance$, for each environment in Tab.~\ref{tab:beta}.

\begin{table}[h!tbp]
    \centering
    \begin{tabular*}{\columnwidth}{l@{\extracolsep{\fill}}rrr}
        \toprule
        Environment & $\beta$ & $AER$ & $\performance$ \\ \midrule
        InvertedPendulum-v$4$ & 1 & $1,000$ & $1$\\
        Hopper-v$4$ & 100 & $1159.6441$ & $0.3249$ \\
        Ant-v$4$ & 1,000 & $1008.5760$ & $0.1830$ \\
        Swimmer-v$4$ & 100 & $107.9770$ & $0.3024$ \\
        HalfCheetah-v$4$ & 1,000 & $147.8970$ & $0.0441$ \\
        \bottomrule
    \end{tabular*}
    \caption{Confidence threshold $\beta$ for MAHALO.}
    \label{tab:beta}
\end{table}

\subsection{Discrepancy Interpretation}

After searching for the best hyperparameter values, we observed a consistent discrepancy in MAHALO's results.
To ensure that this discrepancy was not introduced by the dataset used in our main experiments (IL-Datasets), we also evaluated MAHALO using its original dataset: D$4$RL.
It is important to note that IL-Datasets have been employed in prior studies~\cite{gavenski2024explorative} and consist of $1,000$ episodes generated by trained policies under the same MuJoCo benchmark environments used in most ILfO literature~\cite{torabi2018bco,kidambi2021mobile,zhu2020opolo,gavenski2024explorative}.
Thus, IL-Datasets are not fundamentally different in nature or quality from D$4$RL.
To isolate the impact of the data source, we trained MAHALO on both variants of the D4RL dataset:
\begin{enumerate*}[label=(\roman*)]
    \item using only data collected from a random policy, and
    \item using the full dataset, which includes both random and suboptimal policy trajectories.
\end{enumerate*}

Table~\ref{tab:original_results} reports the results obtained under these settings.
We find that the same performance degradation is present when MAHALO is trained using only random-policy observations, even within the original D$4$RL benchmark. 
In \textit{Hopper-v$2$}, performance decreases by $317.80\%$, and in \textit{HalfCheetah-v$2$}, the drop is more pronounced at $7233.51\%$.
We note that IL-Dataset and D$4$RL have different performances in their teacher datasets.
For \textit{HalfCheetah}, IL-Datasets achieve an average reward of $9,809.94$, while D4RL achieves a slightly higher reward of $10,621.65$.
Similarly, for \textit{Hopper}, IL-Datasets average is $3,531.19$ compared to D$4$RL’s $3,880.89$.

\begin{table}[h!tbp]
    \footnotesize
    \centering
    \begin{tabular*}{\columnwidth}{l@{\extracolsep{\fill}}rr}
        \toprule
        Environment & Random Data & Complete Data \\ \midrule
        Hopper-v$2$ & $1027.47 \pm 0.57$  & $3265.31 \pm 13.95$\\
        HalfCheetah-v$2$ & $64.9124	\pm 5.4846$ & $4695.4462 \pm 1350.801$ \\
        \bottomrule
    \end{tabular*}
    \caption{MAHALO's results for both Hopper and HalfCheetah environments.}
    \label{tab:original_results}
\end{table}

Given these results, we conclude that the observed decrease in performance cannot be attributed to differences in dataset quality, environment version, or seed selection.
Rather, the decline is consistently observed when MAHALO is trained using data from a purely random policy, suggesting that the presence of suboptimal but structured policies in their original setting plays a crucial role in enabling MAHALO's performance.

\end{document}